\documentclass{article}
\usepackage{ijcai17}

\usepackage{times}

\usepackage{amssymb}
\usepackage{color}
\usepackage{graphicx}
\usepackage{tabu}

\newcommand{\pre}{{\textit{Pre}}}

\newcommand{\add}{{\textit{Add}}}
\newcommand{\del}{{\textit{Del}}}

\title{Towards learning domain-independent planning heuristics}
\author{Pawe\l{} Gomoluch\textsuperscript{*}, Dalal Alrajeh\textsuperscript{*}, Alessandra Russo\textsuperscript{*}, Antonio Bucchiarone\textsuperscript{\dag}\\ 
\textsuperscript{*}Imperial College London, UK; \textsuperscript{\dag}Fondazione Bruno Kessler, Trento, Italy\\
\{pawel.gomoluch14,dalal.alrajeh,a.russo\}@imperial.ac.uk, bucchiarone@fbk.eu}

\begin{document}

\maketitle

\begin{abstract}
Automated planning remains one of the most general paradigms in Artificial Intelligence, providing means of solving problems coming from a wide variety of domains. One of the key factors restricting the applicability of planning is its computational complexity resulting from exponentially large search spaces. Heuristic approaches are necessary to solve all but the simplest problems. In this work, we explore the possibility of obtaining domain-independent heuristic functions using machine learning. This is a part of a wider research program whose objective is to improve practical applicability of planning in systems for which the planning domains evolve at run time. The challenge is therefore the learning of (corrections of) domain-independent heuristics that can be reused across different planning domains. 
\end{abstract}

\section{Introduction}

Planning is composing sequences of actions that, when executed starting from an initial state of the environment, will lead to the satisfaction of a given goal. \emph{Classical planning} is a particular type of planning, which relies on the assumption that execution of the actions is the only source of change in the environment, the preconditions and effects of all actions are known and the state of the environment is fully observable. Even under these restrictions, planning is still PSPACE-complete \cite{Bylander1994}. Due to their complexity, planning problems involving nontrivial number of actions and objects require heuristic solutions.

The most successful approaches to classical planning, such as \cite{Hoffmann2001,Helmert2006} are based on heuristic forward search through the space of possible environment states. The core part of a heuristic search algorithm is the \emph{heuristic function}. This is a function that estimates the cost of reaching a goal state from any given state of the environment. By selecting states with relatively low cost estimations, the planner is able to restrict the search to the most promising regions of the search space. Heuristics can be domain-independent and thereby applicable to problems representing various domains, or domain-dependent, in which case they are allowed to rely on knowledge specific to a particular planning domain, possibly encoded by a human expert. Existing domain-independent heuristic functions are hand-crafted and often based on solving simplified (\emph{relaxed}) planning problems \cite{Hoffmann2001,Helmert2006}. On the other hand, heuristics for planning could be learned. But recent attempts in learning planning heuristics are limited to learning knowledge specific to a particular planning domain \cite{Yoon2008,Garrett2016}. 

This paper addresses this gap by providing a first investigation into the extent to which machine learning can be used to automatically learn (corrections of) domain-independent heuristic functions using data generated from different planning domains. Learning domain-independent planning heuristics will help generalize the applicability of planning to a variety of problems including for instance that of dynamic (collective) adaptive systems \cite{Bucchiarone2012}, where the planning domain evolves, together with the system, and thereby is not known a priori. Domain-specific solutions, trained on data representing exactly the same planning domain, would not be applicable in this context. Domain-independence, on the other hand, would allow for learning from large and diverse data sets and having the learner identify patterns occurring across a variety of planning problems and domains.  

It is worth noting that in practice various levels of domain independence are possible. For example, in the aforementioned example of dynamic systems, the different domains may be strongly related: the domain observed at time \emph{t} may typically result from an incremental change to the domain at time \emph{t--1}. The opposite extreme is the case of completely unrelated planning domains, like for example those appearing in the International Planning Competition (IPC) \cite{McDermott2000}. In all these cases, the learning algorithm will need to use a domain-independent representations of the data in order to learn heuristic functions that can be applied to guide the forward search across different planning domains.  In this paper we focus on learning from data generated from completely independent planning domains. This is done using Artificial Neural Networks (ANNs). We then propose an approach for heuristic planning that integrates a (trained) neural network in the planner to guide the forward search. We evaluate the results by comparing the outcomes of solving planning problems using our approach versus existing heuristic forward planners with hard-coded domain-independent heuristics.  

The rest of this paper is structured as follows. Section \ref{related-work} presents related work. Section \ref{background} introduces necessary background. Section \ref{architecture} outlines the architecture of our machine learning-based heuristic planning approach and discusses the underlying learning problem. Section \ref{results} presents the preliminary results obtained on IPC planning domains. Section \ref{conclusion} concludes the paper and suggests directions of our further work.

\section{Related work} \label{related-work}

The idea of improving the planner performance by learning from past plans is almost as old as automated planning itself. The classic STRIPS planner \cite{Fikes1971} was quickly followed by an extension which automatically learned composite actions \cite{Fikes1972}.

However, learning control knowledge for forward search has received limited attention. Notable examples include work by Yoon \emph{et al.} \shortcite{Yoon2007,Yoon2008} who use dedicated learning algorithms to infer domain-specific knowledge in the form of decision lists, \emph{measures of progress} and linear heuristic functions. The learning process leverages features computed based on graphs constructed by the Fast Forward (FF) heuristic \cite{Hoffmann2001}.


Such features are also employed by \cite{Garrett2016}, who use them to learn domain-specific heuristic functions by means of ridge regression or Ranking Support Vector Machines. The application of the latter method is inspired by the fact that, since in forward search heuristic functions are only used to choose between successor states, it may be beneficial to optimize directly for ranking performance of the heuristic rather than accuracy of the estimation of the distance to the goal.

The key difference between our approach and the above works is the aim of domain-independence. To the best of our knowledge, there has been no attempt to learn domain-independent heuristic functions which could be applied in forward planning across a variety of domains.

The aim of learning to plan is inherently related to \emph{reinforcement learning}. One of the key differences between learning-enhanced automated planning and reinforcement learning is the presence of a known and constant model of the environment in the automated planning case. While the necessity to provide such a model is a disadvantage compared to reinforcement settings, the planner's ability to exploit a model enables it to explicitly consider long sequences of steps. This is especially important in \emph{satisficing} planning, where the challenge is to reach any goal state and therefore no reward signal would be available until the problem is solved.

\section{Background} \label{background}

\subsection{Planning}

The aim of classical planning is to compose a sequence of actions which leads to satisfaction of the goal from a given initial state. Formally, a planning task is a tuple $\langle P, O, I, G, A\rangle$, where  $P$ is a set of predicates, $O$ is a set of domain objects, $I$ is the initial state, $G$ is the goal and $A$ is a set of action schemata. The initial state is a set of propositions (instantiations of predicates from $P$ with objects from $O$), which are true before any action is taken. All propositions not included in the set are assumed to be false. An action schema $A$ is a tuple $\langle \pre(A), \add(A), \del(A)\rangle$, where $\pre(A)$ is the set of preconditions -- propositions that must be true before the action is executed, $\add(A)$ is the set of \emph{add} effects -- propositions true after the action is executed, and $\del(A)$ is the set of \emph{delete} effects -- propositions made false by the execution of the action. The state $S(A)$ resulting from execution of action $A$ in state S is defined as follows: $S(A) = S \cup \add(A) \setminus \del(A) $.


For example, the preconditions of driving a vehicle from location A to location B are that the vehicle is indeed in location A and there exists a road between the locations. The add effect  states that the vehicle is now at location B. The delete effect refers to the fact that the car is at location A, which does not hold any more after the action is executed. A simple encoding of the action in PDDL (\emph{Planning Domain Definition Language}, \cite{McDermott2000}) is presented in Figure~\ref{fig:pddl}. The add and delete effects are combined in a single \texttt{and} statement.
\begin{figure}
\begin{verbatim}
(:action drive
  :parameters (?v - vehicle
    ?a ?b - location)
  :precondition (and
    at(?v ?a)
    road(?a ?b)
  )
  :effect (and
    at(?v ?b)
    (not at(?v ?a))
  )
)
\end{verbatim}
\caption{PDDL description of action \emph{drive}.} \label{fig:pddl}
\end{figure}

In planning by \emph{forward search}, sequences of actions are composed by application of concrete planning actions, starting from the initial state. Each action choice yields a new planning state. Since usually many actions are available in a particular state, the search space grows exponentially. For this reason, heuristic functions are used to guide the search towards the goal and allow it to focus on a small subset of the search space. A heuristic function maps any given state-goal pair and returns an estimate of the cost of reaching the goal from the state. The search is then driven towards the goal by preferring the states with low cost estimations. For example, in the \emph{Transport} domain, where packages need to be delivered from their respective origins to destinations, the number of undelivered packages could serve as a very simple heuristic -- the world states in which many packages are already at their destinations have a good chance of being closer to the goal than states for which no packages have been delivered yet.

One of the most successful ideas in heuristic planning is to use a solution of a simplified version of the problem to compute the value of the heuristic. \emph{Delete relaxation} removes the delete effects of the actions, resulting in a much simpler planning problem, whose solution can then be used to designate heuristic values for the original problem \cite{Bonet1997,Hoffmann2001}.

Propositional representation of the environment is convenient for STRIPS-like specification of actions by means of add and delete effects. However, it hides information about relations between the propositions. For example, consider the propositional variables \emph{truck1-at-a} and \emph{truck1-at-b}. Assuming that they refer to the same truck and two distinct locations, they clearly cannot be true both at the same time. This knowledge could be made explicit by using a multi-valued variable \emph{truck1-location} taking values \emph{a}, \emph{b} etc. Such representation is used in so-called \emph{multi-valued planning tasks} and forms the basis of the Fast Downward planning system \cite{Helmert2006}, as well as its Causal Graph heuristic, later generalised to Context-Enhanced Additive heuristic (CEA) \cite{Helmert2008}. The Fast Downward system is extensively applied throughout this work, both to collect data and to test the learned heuristics.

\subsection{Learning}

In this work, we obtain heuristic functions by using supervised machine learning to estimate the cost of reaching the goal from any given planning state: $$ h: S \times G \rightarrow \mathbb{R} $$ where S is the set of possible planning states and G is the set of possible planning goals.

Supervised machine learning is concerned with predicting the value of the \emph{label} (dependent variable) based on the values of the \emph{features} (independent variables). In the particular case when the predicted value is numerical rather than categorical, the learning task becomes a \emph{regression}.

In the simplest setting, the value of the label $y$ can be approximated using a weighted sum of the features $X$. $$ \hat{y} = W^\top X + b$$ This is called \emph{linear regression}. Under some circumstances, a linear model trained with the sole objective of minimizing the error on the training set is likely to \emph{overfit} -- memorize the properties of the training set at the expense of losing generality of the learned function and its ability to correctly label unseen examples. To prevent this effect, \emph{ridge regression} learns a regularized linear model, by minimizing an expression depending not only on the approximation error but also on  $ |W| $, which favours simpler models. Ridge regression is used in this work as the baseline learning algorithm.

To allow for learning more complex, nonlinear functions we use Artificial Neural Networks (ANNs) \cite{McCulloch1943}. ANNs are composed of units called \emph{neurons} which compute a weighted sum of their inputs $I$ and apply a nonlinear activation function $f$ to compute the value of the output $o$. $$ o = f(W^\top I + b) $$ The neurons are arranged in layers with the outputs of one layer serving as the inputs for the next one. Such structures can be trained efficiently by \emph{backpropagation} \cite{Werbos1974} of the prediction error.

In the presented work, we are only concerned with \emph{fully connected} \emph{feed-forward} networks, in which each of the neurons is connected to all the neurons in the previous layer and no output signal is reused as input. Such networks are often referred to as \emph{multi-layer perceptrons}. As the nonlinear function $f$, we use simple rectification \cite{Hahnloser2000}: an identity function truncated at 0 for negative arguments. This is a popular choice for a wide array of ANN applications. Additionally, in our application, a simple activation function is particularly advantageous since it enables very fast state evaluation, with no need for exponentiation operations.

\section{Approach} \label{architecture}

\begin{figure}
\includegraphics[width=\columnwidth]{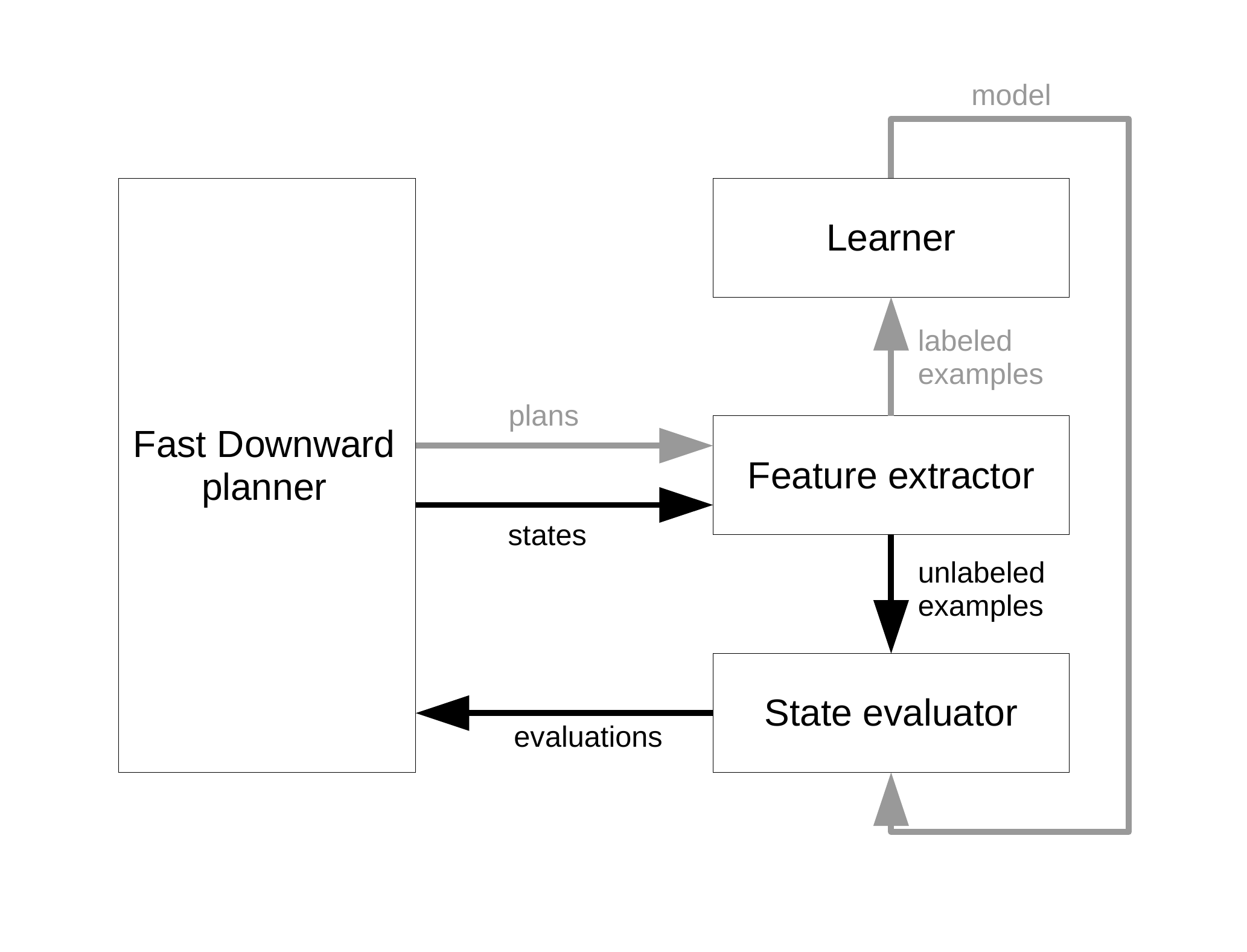}
\caption{Architecture of the proposed solution.} \label{fig:data-flow}
\end{figure}

We propose an approach to heuristic planning based on learning a domain-independent heuristic function. The training data is extracted from known plans for problems representing various domains and used to train a heuristic function in a supervised learning setting. The function is then embedded into the planner and used to evaluate states encountered in the process of solving unseen problems, possibly representing unseen planning domains. In our initial efforts, we restrict experimentation to three domains, all of which are represented in the training data. An overview of the architecture of the approach is presented in Figure \ref{fig:data-flow}. The grey arrows depict the learning data flow. The black arrows show how the system is used to solve new planning problems. The architecture has been implemented as an extension of the Fast Downward planning system. The source code is available online\footnote{https://github.com/pgomoluch/fd-learn}.

\subsection{Data collection}

In this work, the training data comes from solutions to problems from IPC \cite{McDermott2000} domains: \emph{Transport}, \emph{Woodworking} and \emph{Parking}. \emph{Transport} problems are about delivering packages using a fleet of vehicles moving around cities of known topology. The \emph{Parking} domain represents puzzles where a set of cars must be double parked along curbs in their respective slots. In \emph{Woodworking}, wooden parts satisfying various properties need to be engineered from raw wooden boards using a number of tools with different capabilities.

The choice of domains ensures that they are relatively varied. In \emph{Transport}, the goals are relatively independent and once satisfied never need to be undone. This is not the case in \emph{Parking}, where a car may need to be removed from its slot to allow moving another car. \emph{Woodworking}, unlike the two other domains features dead ends - states from which the problem cannot be solved, despite being solvable from the initial state.

For each of the domains, we have generated a set of problems sufficiently small enough to be solved in reasonable time by means of exhaustive search. The actions contained in the solutions have been applied in order, yielding a sequence of states which were labelled with the cost of achieving the goal, computed by subtracting the cost of already executed actions from the overall plan cost. Applied search configuration, first expanding the states nearest to the initial one, guarantees the optimality of obtained solutions and consequently correct cost estimations for all the states on the path to the goal. From the learning point of view, this ensures that there is no noise in the training labels.

For the \emph{Transport} and \emph{Parking} domains, the training problems were generated using the generator scripts provided by the organizers of IPC. In the case of the \emph{Woodworking} domain, the training problems were obtained from one of the demonstration problems, by executing fixed length random walks from the goal state and using the terminal states of the walks as initial states of the new training problems. In total, the data has been extracted from solutions of 280 planning problems. The solutions contained a total of 3956 planning states.

\subsection{Feature extraction}

Representing planning states from various problems and domains in a single feature space is the core challenge of the presented work. Indeed, search for quantities strongly correlated with the distance to the goal is the essence of the research in heuristic planning. At the same time, it is of critical importance to ensure that the features can be computed in reasonable time -- in a degenerate case, feature extraction could leverage the fact that any state-goal pair contains information sufficient to designate the actual distance to the goal -- by optimally solving the planning problem.

In our initial experiments, except for hand-crafting several lightweight features, we follow in the footsteps of \cite{Yoon2008,Garrett2016} and compute features based on existing domain-independent heuristics, including the values of the heuristics themselves. A non-exhaustive list of the features we worked with includes:
\begin{itemize}
\item the number of multi-valued variables in the problem
\item the size of variable domains (number of values a variable can take) -- quartiles of the distribution
\item the number of conjuncts in the goal
\item the number of unsatisfied goal conjuncts (Hamming distance to the goal) -- corresponding to the goal count heuristic
\item the value of the CEA heuristic
\item the value of the FF heuristic
\item the number of delete effects ignored by FF computation of the relaxed plan
\item the number of operators used in the FF relaxed plan.
\end{itemize}

\subsection{Learning}

After initial experimentation with various sets of features, we settled to work with two of them. The first one did not contain any features based on existing heuristics and so required the learner to infer a standalone heuristic by learning from scratch based on very simple features only. The features included information about the number of variables, the size of their domains, the number of the goals and the number of unsatisfied goals.

The second feature set contained information extracted from the computation of the FF heuristic. Except for the value of the heuristic itself, the features included the number of unsatisfied goals, as well as the number of operators used in FF's relaxed plan, the total number of effects ignored by the relaxed plan and the average number of effects ignored. It can be argued that the presence of such advanced features, including the value of another heuristic itself, effectively changes the learning task from learning a standalone heuristic to learning a correction of the base heuristic, in this case FF. Therefore, we further refer to heuristics learned based on this feature set as \emph{FF corrections} and devote particular attention to comparing their performance against the original.

For both  feature sets, two learning approaches have been applied. As a simple baseline approach, regularized linear models were obtained using the ridge regression implementation contained in the scikit-learn library \cite{scikit-learn}. The parameters were then extracted from the model and used to construct a heuristic embedded in the planner. In this setting, the state evaluator component form Figure \ref{fig:data-flow} took the form of an additional Fast Downward heuristic computing the scalar product of the feature and parameter vectors.

To enable learning nonlinear hypotheses, simple neural networks have been trained on the same data.  The networks contained 2 hidden layers composed of rectified linear units and were trained by stochastic gradient descent. The number of neurons in the first hidden layer was equal to the number of features. The second hidden layer contained 3 neurons. The output neuron did not apply nonlinearity to the sum of its inputs. The neural networks have been implemented from scratch. This was to ensure very fast evaluation of single planning states, adding as little overhead as possible to the evaluation time of the heuristic. In this case, the state evaluator component contained the trained network.


\section{Experimental Results} \label{results}

The learned heuristics have been evaluated on three problem sets representing the \emph{Transport}, \emph{Woodworking} and \emph{Parking} domains. In case of \emph{Transport} and \emph{Woodworking}, IPC 14 demonstration sets of 30 problems each were used. For the \emph{Parking} domain, 30 problems have been generated using the problem generator supplied by IPC organizers.

The problems were approached using best-first search guided by various heuristic functions, including FF and CEA (for reference), learned FF corrections and standalone learned heuristics. For every problem, each configuration of the planner was allocated 60 seconds timeout. The experiments were run a desktop computer (Intel i7-4790 CPU, 16 GB RAM). The number of problems solved within the timeout, out of 30 test problems, for every configuration of the planner, is reported in Table \ref{tab:coverage}. In the table, \emph{RR} denotes a heuristic learned using ridge regression and \emph{NN} refers to a heuristic learned using a neural network. \emph{RR-FF} and \emph{NN-FF} are the FF corrections learned using the respective methods. The total number of problems solved within a fixed time constraint is a convenient measure of the relative performance of different heuristics.

In general, heuristics learned with FF-based features (RR-FF and NN-FF, rows 6 and 7 in Table \ref{tab:coverage}) have performed much better than the ones learned from scratch (RR and NN, rows 4 and 5). Good performance of RR and NN heuristics in \emph{Transport} can be attributed to the relative independence of planning goals in this domain. In fact, best-first search guided by the number of unsatisfied goals alone has behaved very similarly to RR and NN, visiting the same numbers of states and overall solving a couple of problems more, thanks to shorter evaluation time of the heuristic. In other words, it turned out that the heuristics learned from scratch essentially replicated the goal count heuristic, equivalent to one of the features used for learning. On the remaining two domains, NN heuristic was the worst, remarkably losing also against the simpler linear model.  

Among FF corrections, the heuristic learned using a neural network (NN-FF, row 7) has performed significantly better than the linear model (RR-FF, row 6) in the \emph{Parking} domain while achieving a similar performance in the two remaining domains. However, it has only slightly outperformed the original heuristic (FF, row 1), solving two more problems in \emph{Transport} and \emph{Parking} but three less in \emph{Woodworking}.

Despite very small improvement over FF in general, the NN-FF configuration turned out to be remarkably better informed than the original heuristic on the first 10 test problems of the parking domain. The number of states generated by the planner guided by the two heuristics is reported in Table \ref{tab:info}. On average, the NN-FF heuristic required generation of 3.5 times less states than FF (the ratios were aggregated using geometric mean). However, when averaged all the parking problems solved by both of the heuristics, the ratio fell to 1.32, indicating that the advantage gained by NN-FF on easier problems is lost on the harder ones.


\begin{table}
\centering
\begin{tabu}{c|ccc}
& Transport & Woodworking & Parking \\\hline
FF & 10 & 26 & 19 \\
CEA & 20 & 23 & 20 \\
Goal count & 22 & 14 & 14 \\
RR & 22 & 13 & 11 \\
NN & 18 & 6 & 3 \\ 
RR-FF & 12 & 23 & 5 \\
NN-FF & 12 & 23 & 21\\
\end{tabu}
\caption{Number of problems solved (out of 30 for each domain)} \label{tab:coverage}
\end{table}

\begin{table*}
\centering
\begin{tabu}{c|cccccccccc}
& 1 & 2 & 3 & 4 & 5 & 6 & 7 & 8 & 9 & 10 \\\hline
FF & 34 & 153 & 1194 & 818 & 1876 & 2207 & 4068 & 13671 & 5232 & 42964 \\
NN-FF & 34 & 52 & 482 & 260 & 470 & 631 & 2228 & 1198 & 2428 & 2191 \\
\end{tabu}
\caption{Number of states generated for the first 10 problems of the parking domain} \label{tab:info}
\end{table*}

\section{Conclusions and Further Work} \label{conclusion}

In this paper, we have presented our first attempt at learning domain-independent planning heuristics from optimal solutions to small planning problems. On average, the heuristic have not performed better than state of the art domain-independent heuristics, even in cases when the values of the heuristics were included in the feature set, effectively changing the learning task from learning a standalone heuristic function to learning a correction of an existing heuristic.

However, the FF correction obtained using a neural network has shown a promising property of being significantly better informed than the original heuristic on a specific subset of the test problems.

We are currently working on improving the learning approach in order to extend the range of planning domains and problems on which it offers an improvement over the original heuristic. Our efforts are primarily focused on constructing a better feature representation of the planning states. As one of the possible directions, we are considering augmenting the feature set with more features related to the graphs constructed by FF and CEA.

Moreover, we plan to extend the method of data collection in order to acquire data representative of larger planning problems. Currently, the size of the training problems is severely restricted because their solutions are computed using exhaustive search. In the future, the training data will also contain high-quality heuristic solutions of larger problems. The solutions will be obtained by selecting the lowest-cost plans among those computed by various heuristic planners, operating at large timeouts and with relatively conservative search routines. Since the solutions will no longer be optimal, this way of collecting plans will inevitably introduce noise to the dataset. However, the drop in relative performance of the NN-FF heuristic on harder problems suggests that ensuring some representation of larger cases may be the key to further improvement.

\bibliographystyle{named}
\bibliography{ijcai17}

\begin{thebibliography}{}

\bibitem[\protect\citeauthoryear{Bonet \bgroup \em et al.\egroup
  }{1997}]{Bonet1997}
Blai Bonet, Gabor Loerincs, and Hector Geffner.
\newblock {A Robust and Fast Action Selection Mechanism for Planning}.
\newblock {\em Proceedings of the fourteenth national conference on artificial
  intelligence and ninth conference on Innovative applications of artificial
  intelligence}, pages 714--719, 1997.

\bibitem[\protect\citeauthoryear{Bucchiarone \bgroup \em et al.\egroup
  }{2012}]{Bucchiarone2012}
Antonio Bucchiarone, Annapaola Marconi, Marco Pistore, and Heorhi Raik.
\newblock {Dynamic Adaptation of Fragment-based and Context-aware Business
  Processes}.
\newblock In {\em 2012 IEEE 19th International Conference on Web Services, ICWS
  2012}, pages 33--41, 2012.

\bibitem[\protect\citeauthoryear{Bylander}{1994}]{Bylander1994}
Tom Bylander.
\newblock The computational complexity of propositional strips planning.
\newblock {\em Artif. Intell.}, 69(1-2):165--204, 1994.

\bibitem[\protect\citeauthoryear{Fikes and Nilsson}{1971}]{Fikes1971}
Richard~E Fikes and Nils~J Nilsson.
\newblock {STRIPS: A New Approach to the Application of Theorem Proving to
  Problem Solving}.
\newblock {\em Artificial Intellegence}, 2:189 -- 208, 1971.

\bibitem[\protect\citeauthoryear{Fikes \bgroup \em et al.\egroup
  }{1972}]{Fikes1972}
Richard~E Fikes, Peter~E Hart, and Nils~J Nilsson.
\newblock {Learning and Executing Generalized Robot Plans}.
\newblock {\em Artificial Intelligence}, 3(1972):251--288, 1972.

\bibitem[\protect\citeauthoryear{Garrett \bgroup \em et al.\egroup
  }{2016}]{Garrett2016}
Caelan~Reed Garrett, Leslie~Pack Kaelbling, and Tom{\'{a}}s
  Lozano{-}P{\'{e}}rez.
\newblock {Learning to Rank for Synthesizing Planning Heuristics}.
\newblock In {\em Proceedings of the Twenty-Fifth International Joint
  Conference on Artificial Intelligence, {IJCAI} 2016}, pages 3089--3095, 2016.

\bibitem[\protect\citeauthoryear{Hahnloser \bgroup \em et al.\egroup
  }{2000}]{Hahnloser2000}
R~H Hahnloser, R~Sarpeshkar, M~A Mahowald, R~J Douglas, and H~S Seung.
\newblock {Digital selection and analogue amplification coexist in a
  cortex-inspired silicon circuit.}
\newblock {\em Nature}, 405(6789):947--951, 2000.

\bibitem[\protect\citeauthoryear{Helmert and Geffner}{2008}]{Helmert2008}
Malte Helmert and Hector Geffner.
\newblock Unifying the causal graph and additive heuristics.
\newblock In {\em Proceedings of the Eighteenth International Conference on
  Automated Planning and Scheduling, {ICAPS} 2008}, 2008.

\bibitem[\protect\citeauthoryear{Helmert}{2006}]{Helmert2006}
Malte Helmert.
\newblock {The Fast Downward Planning System}.
\newblock {\em Journal of Artificial Intelligence Research}, 26:191--246, 2006.

\bibitem[\protect\citeauthoryear{Hoffmann and Nebel}{2001}]{Hoffmann2001}
J{\"{o}}rg Hoffmann and Bernhard Nebel.
\newblock {The FF Planning System: Fast Plan Generation Through Heuristic
  Search}.
\newblock {\em Journal of Artificial Intelligence Research}, 14:263--312, 2001.

\bibitem[\protect\citeauthoryear{McCulloch and Pitts}{1943}]{McCulloch1943}
Warren~S. McCulloch and Walter Pitts.
\newblock A logical calculus of the ideas immanent in nervous activity.
\newblock {\em The Bulletin of Mathematical Biophysics}, 5(4):115--133, 1943.

\bibitem[\protect\citeauthoryear{McDermott}{2000}]{McDermott2000}
D.~McDermott.
\newblock {The 1998 AI planning systems competition}.
\newblock {\em AI magazine}, 21(2):1--33, 2000.

\bibitem[\protect\citeauthoryear{Pedregosa \bgroup \em et al.\egroup
  }{2011}]{scikit-learn}
F.~Pedregosa, G.~Varoquaux, A.~Gramfort, V.~Michel, B.~Thirion, O.~Grisel,
  M.~Blondel, P.~Prettenhofer, R.~Weiss, V.~Dubourg, J.~Vanderplas, A.~Passos,
  D.~Cournapeau, M.~Brucher, M.~Perrot, and E.~Duchesnay.
\newblock Scikit-learn: Machine learning in {P}ython.
\newblock {\em Journal of Machine Learning Research}, 12:2825--2830, 2011.

\bibitem[\protect\citeauthoryear{Werbos}{1974}]{Werbos1974}
P.~J. Werbos.
\newblock {\em Beyond Regression: New Tools for Prediction and Analysis in the
  Behavioral Sciences}.
\newblock PhD thesis, Harvard University, 1974.
\newblock Department of Applied Mathematics.

\bibitem[\protect\citeauthoryear{Yoon \bgroup \em et al.\egroup
  }{2007}]{Yoon2007}
Sung~Wook Yoon, Alan Fern, and Robert Givan.
\newblock {Using Learned Policies in Heuristic-Search Planning}.
\newblock In {\em IJCAI 2007, Proceedings of the 20th International Joint
  Conference on Artificial Intelligence}, pages 2047--2053, 2007.

\bibitem[\protect\citeauthoryear{Yoon \bgroup \em et al.\egroup
  }{2008}]{Yoon2008}
Sung~Wook Yoon, Alan Fern, and Robert Givan.
\newblock {Learning Control Knowledge for Forward Search Planning}.
\newblock {\em The Journal of Machine Learning Research}, 9:683--718, 2008.

\end{thebibliography}

\end{document}